# From LLM to LDM: Large Discourse Models and the Governance of Artificial Discursive Agents

*Revised and expanded English-language version. This text constitutes an independent contribution derived from, but not a translation of, earlier work in French.*

Amar LAKEL
Associate Professor
Bordeaux Montaigne University
MICA Laboratory
amar.lakel@u-bordeaux-montaigne.fr

**Abstract**

This article proposes a decisive conceptual reframing: the redefinition of "Large Language Models" as "Large Discourse Models" (LDMs) and the establishment of a new analytical category, the Artificial Discursive Agent (ADA). Drawing on a semiotico-pragmatic theoretical framework that distinguishes three regulatory instances — phenomenal regularities, embodied cognition, and discursive sedimentation — we argue that these systems must be evaluated not at the level of linguistic competence but at the level of their discursive practices: the reproduction of genres, enunciative positions, argumentative framings, and normative alignments as sedimented in their training corpora. We trace the genealogy of algorithmic models in the social sciences from Quetelet's social physics through Benzecri's correspondence analysis to contemporary deep learning architectures, situating the emergence of LDMs within this trajectory. We propose four cumulative evaluation criteria for ADAs (discursive generalization, attributable rational coherence, normative alignment, traceable biography) and a falsifiable empirical program of five hypotheses (P1–P5) designed to test their capacities, measure their socialization effects — including cognitive debt, competence compression, and educational dependency — and establish the governance procedures through which democratic societies can determine their place, uses, and limits.

## INTRODUCTION

The rise of large generative models marks a significant shift in digital studies research. The field is moving from the analysis of abstract language processing toward an examination of computable discursivity. In this paper, we investigate how these systems challenge established interpretive practices and transform the methods, objects, and validation frameworks through which scholars assess utterances. We contend that addressing these questions requires reconnecting with the theoretical heritage and hermeneutic traditions of the humanities and social sciences.

The dominant critical framework for evaluating large language models (the "stochastic parrots" thesis advanced by Bender and Koller (2020) and Bender et al. (2021)) holds that systems trained exclusively on textual data manipulate linguistic form without access to meaning, and therefore cannot be said to understand language in any substantive sense. Let us grant the thesis its maximal extension. Let us concede that these systems do not understand in the way that embodied human subjects understand, that they lack phenomenological grounding, intentional consciousness, and subjective experience. But the concession amounts to a tautology: an artificial intelligence is not a human intelligence. No serious theoretical position has ever claimed otherwise. The problem is that this tautology, however accurate, exhausts the analytical capacity of the framework that produces it. By demonstrating that these systems are not human minds, the stochastic parrots thesis forecloses the only question that now demands investigation: what kind of intelligence *is* this, and what place should it occupy within the social arrangements through which human communities organize their collective existence?

The framework offers no gradient between "mere mimicry" and "genuine comprehension," no analytical vocabulary for characterizing the specific modality of cognitive-discursive competence that these systems demonstrably exhibit, and no instruments for determining — through public, testable procedures — the conditions under which such competence warrants integration into, or exclusion from, the institutions of democratic life. It predicts nothing about the emergent capacities observed since 2022 (Piantadosi & Hill, 2022; Li et al., 2023; Arora & Goyal, 2023; Huh et al., 2024), explains nothing about the mechanisms through which distributional learning produces quasi-rule-governed discursive behavior, and provides no basis for the governance decisions that democratic societies must now make. The stochastic parrots thesis identifies a real boundary (the boundary between human and artificial cognition). But it mistakes the identification of this boundary for the resolution of the problem, when in fact the problem begins precisely where the boundary is drawn: on the other side stands an intelligence of a new kind, already operating within social fields, already producing discursive performances that human communities receive, evaluate, and act upon, and whose characterization requires conceptual instruments that the binary vocabulary of the grounding debate cannot supply.

Piantadosi and Hill (2022) relocated meaning from referential grounding to conceptual role semantics. Li et al. (2023) proved experimentally that next-token prediction generates emergent world models. Arora and Goyal (2023) demonstrated compositional generalization beyond the training corpus. Huh, Cheung, Wang, and Isola (2024) showed that neural networks converge toward a shared statistical model of reality across modalities, architectures, and objectives. Hinton (2023) stated the obvious: accurate prediction at scale requires understanding. They prove that something more than parroting is occurring; they cannot name what it is. None of them disposes of the conceptual apparatus to characterize what these systems model when they model human discourse, not a neutral "reality" but a politically structured, institutionally stratified, ideologically saturated discursive order. And none of them, consequently, can pose the question that the governance of these systems now demands: not whether Large Discourse

Models "understand," but whose discursive positions they reproduce, whose they suppress, and through what procedures these operations can be identified, audited, and regulated. The framework we develop here — the ontological sorting of regulatory instances, the category of Large Discourse Models, and the concept of Artificial Discursive Agents — provides the instruments to move from the epistemological debate to the operational one.

We propose a decisive reframing. The stochastic parrots thesis fails because it misidentifies what the training corpus contains — and it misidentifies the corpus because it reads it through an epistemology of language (the individual subject and her referential relation to the world) rather than through a theory of discourse (the socially constituted formations that determine what can be said, by whom, and under what conditions). A corpus of billions of documents produced by millions of embodied, cognitively structured, socially situated human speakers is not "form without meaning." It is the documentary sedimentation of a discursive order — a compressed, reified, but structurally faithful encoding of how a civilization perceives reality, constructs logical relations, distributes normative positions, and regulates the conditions of legitimate enunciation. By learning to predict the next token across this corpus, Large Discourse Models have not merely memorized surface patterns. They have internalized a functional model of the discursive conditions under which humans produce interpretable utterances — a model sufficiently powerful that these systems now win the Turing test against human judges, sustain coherent argumentative positions across thousands of tokens, adapt fluidly to genre, register, and enunciative constraints, and exhibit cross-domain transfer capacities that no theory of "mere form manipulation" can account for.

The imitation, if it is one, is not shallow. Whether or not these systems "understand" in any sense that would satisfy the criteria of embodied cognition (Varela, Thompson & Rosch, 1991), the discursive performances they produce are not reducible to lexical recombination: they reproduce the deep structures — genres, argumentative framings, enunciative positions, normative alignments — through which human communities construct, contest, and transmit their experience of the world. This observation warrants neither celebration nor dismissal. It warrants a precautionary posture. If what these systems approximate is not language as formal code but discourse as civilizational practice, then the risk they pose is not the risk of a tool that malfunctions — it is the risk of an agent that functions, that functions at the level of the discursive formations through which social order is produced and legitimated, and whose functioning we do not yet possess the instruments to evaluate at the appropriate analytical level. And if the trajectory of these systems — from task-specific performance to cross-domain discursive generalization — continues along the lines that recent developments suggest (Brown et al., 2020; Bubeck et al., 2023), the possibility must be considered that they will produce synthetic discursive positions of a scope that no individual human speaker could occupy, aggregating perspectives, registers, and normative orientations across domains at a scale that exceeds the finitude of any single embodied life. This possibility may prove illusory. But the cost of assuming it illusory without investigation is asymmetrically higher than the cost of developing the analytical instruments required to characterize it. The theoretical framework we develop in this paper — the ontological sorting of regulatory instances, the category of Large Discourse Models, and the concept of Artificial Discursive Agents — provides these instruments.

The argument proceeds in three parts. Part One establishes the theoretical framework: the ontological sorting that distinguishes three regulatory instances — phenomenal, cognitive, and discursive — and demonstrates that LDMs operate exclusively at the third level, accessing the first two only through documentary mediation. Part Two develops the category of Artificial Discursive Agent (ADA), defined not by what the system "is" but by what it demonstrably does, and proposes four cumulative evaluation criteria — discursive generalization (C1), attributable rational coherence (C2), normative alignment (C3), and traceable biography (C4) — designed to assess these systems at the discursive level rather than at the linguistic or computational level where current benchmarks operate. Part Three examines the conditions under which democratic societies can govern these agents through what we term "polis tests" — public evaluation procedures administered by situated communities of judgment — and articulates the institutional framework necessary to render decidable the place, uses, and limits of artificial discursive agents within the public sphere.

## THEORETICAL FRAMEWORK: FROM SIGNIFICATION TO ONTOLOGICAL SORTING

### THE DOCUMENT AS SEMIOTIC PROJECTION: THREE REGULATORY INSTANCES

We ground our position in a semiotico-pragmatic definition of discourse. On this account, meaning — what we term *signification* — emerges from practice that is both anchored and eventful. Following Foucault, we locate meaning "neither in words nor in things" (1969, p. 66). The signifying structure articulates two dimensions. On one side stand constraining regularities: institutional rules, grammars, and explicit norms. On the other stand social usages that communities have stabilized through situated practice. Within this structure, the speaker actualizes an occurrence that is at once unique and constrained — an event irreducible to individual initiative yet inseparable from the regulatory forces that made it possible.

The utterance, in its documentary reification, constitutes an interpretable artifact. This artifact instantiates a *discursive position* (Guilhaumou, 2005; Pêcheux, 1969; Robin, 1973) in which several elements converge: the speaker's *ethos* — the discursive construction of credibility and character within the utterance itself (Maingueneau, 2002) —, the genre — the relatively stable types of utterances organizing communication within particular spheres of activity (Bakhtine, 1984) —, the referent, the communicational situation, and the address (Jakobson, 1963; Charaudeau, 2004). Speech act pragmatics (Austin, 1962; Searle, 1969), articulated with the linguistics of enunciation (Benveniste, 1966; Ducrot, 1984), establishes that every utterance simultaneously constitutes a saying and a doing: it accomplishes an action within a communicative context. The document is not a transparent vehicle for pre-existing intentions. Marked by a communicative situation, it inscribes itself within what Foucault terms *discursive formations* — sets of utterances governed by common rules of formation, irreducible to the initiative of individual speakers. Foucault (1971) emphasizes within these formations the *procedures of discourse control* through which societies regulate what can be said, by whom, and under what circumstances. The document reifies a meaning constructed for another within a space of constraints that both precedes and exceeds it.

The interpretation of documents — including algorithmic interpretation — thus articulates three distinct regulatory instances, each constituting a class of constraints that weighs upon the genesis of the utterance and leaves identifiable traces within it.

(1) The first instance consists of *phenomenal regularities* of the referential world — perception and action — as inscribed in the speaker's experience. The medieval nominalist tradition, from Ockham to Buridan, established a foundational claim: the general categories through which we organize experience are constructions of the intellect that signify particulars, not entities subsisting independently of the mind (Adams, 1987). This orientation does not collapse into skepticism; it maintains cognitive access to particulars while foregrounding the constructed, parsimonious character of our classificatory systems. Contemporary cognitive science extends this insight: within the Bayesian processing paradigm, the cognitive system operates as an active translation instance, continuously generating hypotheses about the causes of its sensory states and revising them against ascending signals (Clark, 2013; Friston, 2010). The organism does not passively receive data. It anticipates environmental regularities through hierarchical hypothetical models learned from experience, encoding conditional probability distributions that categorize sensory configurations along a small number of distinctive features. The referential instance thus concerns the grip of real-world experience upon the utterance: indexical traces — deictics, proper names, definite descriptions — anchoring discourse in a world of reference whose regularities the speaker has internalized through embodied interaction.

(2) The second instance consists of **regularities of embodied cognition** — mind and body — constituting the speaker as subject. Language constitutes an additional stratum of transformation — a translational overlay upon inner thought (Tomasello, 2003; Varela, Thompson & Rosch, 1991). The utterance emerges as a translation of a translation: the conversion of cognitive experience into a communicable artifact. This poietic capacity renders the document a meaningful object detachable from its context of production, yet one that retains the trace of the cognitive structures that shaped it. The subjective instance concerns the enunciative position: discursive *ethos*, modalizations marking the speaker's degree of adherence to her own utterances, axiological isotopies betraying an evaluative orientation (Rastier, 2009). The speaker does not stand before language as a user before a tool; the speaker is constituted as speaker within and through cognitive-linguistic structures that organize her relation to the world, to others, and to her own discourse.

(3) The third instance consists of **discursive regularities** governing the conditions of utterance production as sedimented in a corpus of situated documents. These are the regularities that Foucault's archaeology (1969) identifies as historically constituted rules of formation governing what can be said, and that Pêcheux's (1969, 1975) materialist discourse analysis links to the ideological formations structuring social relations of production. The communicative instance concerns the apparatus of address: discursive genre, communicative contract (Charaudeau, 2004), pragmatic horizon of action. The exegetical approach developed within French automated discourse analysis (Pêcheux, 1969; Maingueneau, 1984; Robin, 1973) instruments interpretation through the objectification of recurrences, posing reading hypotheses that are revisable through confrontation with textual regularities. The correlation between intra-discursive patterns, inter-discursive regularities, and extra-discursive observations (historical context, sociological data, archives) makes it possible to evaluate the presence of the three regulatory instances without claiming to reach a truth of the world: exegesis constructs testable correspondences between discursive forms and hypotheses concerning their conditions of production.

Every document in a corpus thus constitutes a semiotic projection of these three regulatory instances — a surface in which phenomenal constraints, cognitive structures, and discursive formations have left their respective traces through the act of enunciation. The document does not contain the referential world, the embodied mind, or the institutional formation that produced it. It contains their *projections* — compressed, reified, partially deformed, but structurally identifiable through the methods that the humanities and social sciences have developed across the hermeneutic, empiricist, and critical traditions. The ambition of these disciplines has always been to explore the corpus as a sociohistorical instance informing practices of production and interpretation. In pursuing this aim, they equipped themselves, by the end of the twentieth century, with a techno-methodological apparatus — textual statistics, network analysis, classification algorithms — that objectifies discursive regularities into formalized representations.

It is within this trajectory that the emergence of large generative models must be situated. Our central theoretical claim is that these models constitute the most powerful formal imprint of the algorithmic hermeneutic process to date: through their billions of parameters, neural networks have found a way to compute, at unprecedented scale, the regularities sedimented in the documentary projections of the three regulatory instances. They model the discursive positions of a population as that population has made itself available for interpretation in a corpus. Like artificial archaeologists, they interpret the semiotic projection of a human world whose direct experience they lack. It is at this pragmatico-discursive level — not at the level of natural language understanding, not at the level of referential grounding — that they must be characterized and evaluated, both in their power and in their limits.

## THE REDUCTIONIST MOMENT AND ITS EPISTEMOLOGICAL LIMIT

The computational formalization of language in the second half of the twentieth century constituted a necessary scientific moment — the reductionist translation of discursive phenomena into manipulable formal structures. Chomsky's (1965) generative grammar, the symbolic AI that followed (formal grammars, expert systems, rule-based NLP), and Pêcheux's (1969) first automated discourse analysis all participated in this enterprise. Reductionism is not error; it is a condition of scientific intelligibility — the controlled simplification through which a domain of phenomena becomes tractable (Bachelard, 1938). Every model is, in Latour's (1999) terms, a translation that is simultaneously a betrayal: it captures certain regularities at the cost of suppressing others. The question is never whether the model reduces — all models reduce — but whether what it suppresses is marginal or constitutive.

What symbolic formalization suppressed was constitutive. The generative program treated discourse as the product of language — the output of a formal competence possessed by an idealized speaker-hearer. But language, in the sense of a morphosyntactic system governed by finite rules, constitutes only one regulatory instance among those that determine the production of utterances. The body, perception, affective states, institutional positions, power relations, genre constraints, historical conjunctures, and the irreducible contingency of communicative situations all preside over discourse (Foucault, 1969; Bourdieu, 1982; Bakhtine, 1984). The utterance is not generated by a grammar; it emerges from the intersection of forces — linguistic, cognitive, social, political, material — whose configuration exceeds any formal system designed to capture language alone. This is precisely what the "AI winter" revealed, read at the epistemological rather than the technical level: not a shortage of computing power but a fundamental mismatch between the level at which formalization operated (language) and the level at which meaning is constituted (discourse).

The recognition that meaning is historically constituted rather than formally derivable defines an epistemological tradition that runs from Bachelard's (1938) historical epistemology through Kuhn's (1962) paradigm theory and Wittgenstein's (1953) analysis of language games to Foucault's (1969) archaeology of discursive formations. These thinkers share a common insight: the conditions under which statements acquire meaning, validity, and force are not logical invariants but historical configurations (*epistemes*, paradigms, forms of life) that must be described in their specificity rather than deduced from formal principles. Foucault is a historian of discourse for the same reason that Bachelard is a historian of science and Kuhn a historian of paradigms: because the rules governing what can be said, thought, and known are not transcendental structures but contingent formations produced by practices that are at once cognitive, institutional, and political.

The distributional tradition provided the first computational instruments capable of operating at this level of complexity without requiring explicit rule specification. Harris (1954) proposed a linguistics grounded in co-occurrence regularities observable in corpora, without recourse to the postulate of an *a priori* generative instance. Benzécri (1973) generalized this approach through correspondence analysis: meaning becomes a positional effect within a multidimensional space, not the decoding of a pre-established signified. Bourdieu (1979) demonstrated its sociological power in *Distinction*. The trajectory from distributional statistics through Latent Semantic Analysis (Landauer & Dumais, 1997), Latent Dirichlet Allocation (Blei, Ng & Jordan, 2003), dense vector representations (Mikolov et al., 2013; Pennington et al., 2014), and contextualized embeddings (Peters et al., 2018) to the Transformer architecture (Vaswani et al., 2017) and the GPT family (Radford et al., 2018; Brown et al., 2020) extends this genealogy while effecting a rupture. Probing experiments confirm a hierarchical organization of neural network layers from syntax through semantics to discourse-level phenomena — coherence, rhetorical relations, coreference (Tenney et al., 2019; Koto et al., 2021). If deep layers encode discourse coherence and rhetorical structure, they necessarily encode — at least implicitly — the genre conventions, register norms, and enunciative positions that govern them.

But this achievement was obtained at a price that constitutes the central epistemological problem of the present moment. At every stage of this trajectory, a consistent exchange operated: the gain in formalization capacity was purchased through a corresponding loss of interpretive transparency. Benzécri's factorial axes could be read and named by the analyst; the billions of parameters of a Transformer-based model constitute a "black box" (Pasquale, 2015) whose internal representations escape expert analysis. It was necessary to abandon analysis — the decomposition of phenomena into interpretable components — to a process more powerful than the human analyst in order to access a model capable of imitating the full complexity of discursive production. The symbolic approaches reached their limit because discourse exceeds language; the distributional approaches succeeded because they operated at the level of discourse without requiring its prior decomposition into explicit rules. The condition of this success was the progressive evacuation of interpretive comprehension from the modeling process. The perfect imitative model — the LDM capable of reproducing genres, enunciative positions, argumentative framings, and normative alignments — is obtained at the cost of what we term the *alienation of interpretation* : the transfer of an interpretive capacity to an external instance that operates it while rendering its operations opaque to the subject who depends upon them. In the classical paradigm (Quetelet, Benzécri, Bourdieu), the model was an instrument of comprehension; the analyst retained interpretive sovereignty. When the model becomes a black box whose discursive performances exceed the analyst's capacity to explain them, interpretive sovereignty migrates from the human subject to the computational system — or, more precisely, to the corporations that build, train, and deploy these systems under conditions that are themselves opaque (Crawford, 2021; Zuboff, 2019; Srnicek, 2017). This double alienation — epistemological and political — is what makes the evaluative framework developed in the following sections not merely analytically productive but democratically urgent.

## EMPIRICAL PROGRAM: ADAPTING TESTS AND METRICS TO ARTIFICIAL DISCURSIVE AGENTS

The theoretical framework established in Part One defines the epistemological situation within which any empirical program must now operate. We do not possess a transparent theory of what LDMs have formalized in their billions of parameters; neural network interpretability is advancing (Tenney et al., 2019; Elhage et al., 2022) but has not yet rendered legible the discursive operations these systems perform. The question is therefore: what can be known, evaluated, and governed under conditions of constitutive opacity? Where the public debate oscillates between uncritical enthusiasm (Chui et al., 2023; Eloundou et al., 2023) and existential alarm (Bostrom, 2014; Bengio et al., 2024), we propose a third path: the methodical construction of the instruments through which democratic societies can determine, on empirical grounds, what place these agents should occupy within their institutions. This program is organized around three axes.

(1) The first consists of characterizing the genealogy of LDMs — documenting their biographies (corpora, architectures, alignment procedures, inter-version deltas) in order to constitute the descriptive infrastructure for a future interpretive science of these systems.
(2) The second consists of evaluating the discursive functions of Artificial Discursive Agents as situated agents within social and community networks, through the four criteria (C1--C4) established above.
(3) The third consists of designing cumulative protocols for assessing the impacts and conditions of socialization of ADAs within the institutions — educational, professional, deliberative — where their deployment carries direct consequences for cognitive autonomy and democratic governance.

## From LLM to LDM: A Terminological Displacement and Its Epistemological Stakes

The analysis developed thus far compels a terminological displacement whose epistemological stakes we now intend to establish. We propose replacing the category "Large Language Models" (LLM) with that of Large Discourse Models (LDM). The training corpus constitutes the documentary reduction of a collective expression of human experience situated in space and time. The more this corpus expands and diversifies, the greater the fraction of a documented culture's discursive regularities the model captures. Human language is a system of symbolic compression of experience that retains the structural trace of regularities in the lived world[1]. Training corpora, condensing billions of utterances produced by embodied cognitive subjects, carry the statistical imprints of shared mental models: verbalized causal relations, action scripts, perceptual categorizations, and normative judgments crystallize in textual distributions. This analysis establishes three articulated ontological levels: the phenomenal world; the embodied human cognition that models it; and the discursive sedimentation of that experience in documentary production. The LDM operates exclusively at the third level, accessing the first two only through documentary mediation.

[1]Michael Tomasello, *Constructing a Language: A Usage-Based Theory of Language Acquisition*, Cambridge, Harvard University Press, 2003 ; Andy Clark, *Being There: Putting Brain, Body, and World Together Again*, Cambridge, MIT Press, 1997

The transformations of the semantic world effected by LDMs thus depend on four evolutionary axes whose cartography constitutes a methodological prerequisite for evaluation :

*Table 1 — Evolutionary Axes of Large Discourse Models*

| Axe | Description |
|---|---|
| A1. Corpus and documentary experience | Scope, quality, diversity, and debiasing of training and contextual enrichment data. Access to the corpus conditions the emergence of capabilities. Three distinct modalities of documentary incorporation operate at different temporal scales: initial training, fine-tuning, and context engineering (RAG, knowledge graphs, extended context windows). |
| A2. Neural architectures | Innovations in algorithmic structures: depth, attention mechanisms, integration of external tools, multimodal architectures. The proliferation of variants makes each model a singular configuration within a rapidly expanding ecosystem. |
| A3. Normative education and alignment | Integration of behavioral norms through documented protocols (RLHF, RLAIF, Constitutional AI). Alignment, intrinsically social and cultural, requires the learning of usage conventions specific to each community. The reinforcement protocol determines the acceptability of outputs within situated contexts. |
| A4. Interactional memory | Session context (prompt, conversational history) shapes performance. LDMs are embedded within software architectures integrating search functions, persistent instructions, and access to external resources — configurations we designate as Artificial Discursive Agents (ADA) (see below, Empirical Program). |

These four axes combine into singular configurations — the models — defining the developmental state of a given ADA. Their systematic characterization constitutes a major methodological challenge for any empirical research on these discursive agents of a new kind.

The systematic population of this grid confronts, however, an informational asymmetry that is constitutive rather than accidental: for proprietary models (GPT-4, Claude, Gemini), the training corpora are undisclosed, the architectural specifications redacted, the alignment procedures described without reproducible detail, and the inter-version modifications implemented without public documentation. The researcher confronts an object whose behavior she can observe but whose constitutive factors she cannot inspect. The situation is analogous to the one that the nascent social sciences faced in the nineteenth century: when Quetelet (1835) undertook his "social physics" and Durkheim (1897) sought to correlate suicide rates with social variables, the statistical infrastructure necessary to describe populations did not exist — it had to be constructed through census protocols, standardized categories, and institutional agreements on what would count as a datum (Desrosières, 1993). The *social science of artificial discursive agents* finds itself in a comparable foundational moment. The protocol must therefore be designed to produce cumulative knowledge under conditions of partial information: at minimum, the dated model version, its observable behavioral characteristics (language coverage, refusal patterns, stylistic defaults), its accessible context window, and its deployment conditions — what we term the *behavioral phenotype* of the agent, documented through standardized evaluation batteries administered under controlled parameters. For open-source and open-weight models (OLMo, BLOOM, Llama), the four axes can be characterized with a precision that permits inferential — rather than merely correlational — analysis of the relationship between constitutive factors and discursive behavior; the open-source ecosystem thus constitutes the natural laboratory where the descriptive protocol can be developed and calibrated before being applied, in its necessarily degraded form, to the proprietary systems that dominate actual deployment. The progressive enrichment of this nomenclature — as interpretability science advances, as regulatory pressure compels greater disclosure (EU AI Act, 2024), and as the open-source ecosystem expands — will gradually transform behavioral phenotyping into inferential audit. But the foundational operation must begin now, with whatever information is available, because the alternative is not methodological caution but empirical blindness before an object that is already reshaping the discursive conditions of social life.

## A New Intelligent Species? The Artificial Discursive Agent (ADA)

The concept of Artificial Discursive Agent rests on an epistemological displacement that articulates three operations : the adoption of an anti-essentialist problematic that defines the social status of the artificial agent not by what it "is" but by what it demonstrably does: public, testable, functional criteria of enunciative behavior assessed by situated communities of judgment. This approach is rooted in the pragmatic tradition of Dennett's intentional stance (2009), which evaluates agents by their behavioral outputs rather than their intrinsic properties, extended by Chalmers' computational philosophy of mind (2011) and, in the ANT tradition, by Latour's symmetrical treatment of human and non-human agents within sociotechnical assemblages (2005) and Callon's sociology of translation (Akrich, Callon & Latour, 2006).

The ADA constitutes a classificatory and regulatory sociotechnical category organizing the evaluation, deployment, and accountability of a source of behaviors to which the qualifier "intelligent" cannot be denied without anthropocentric discrimination. This claim requires clarification. We do not argue that ADAs possess consciousness, intentionality, or subjective experience. We argue that the refusal to evaluate their cognitive-discursive competences on their own terms (the insistence that only biologically grounded cognition can qualify as "intelligent") constitutes a disciplinary prejudice rather than a scientific position. ADAs manifest cognitive-discursive competences that merit evaluation according to a dedicated assessment framework. We propose four cumulative criteria characterizing their discursive agency[2]:

**C1 — Discursive generalization and robustness.** The capacity to learn and generalize discursive regularities — logical, stylistic, and ethical — from a documentary base, manifested by the projection of rhetorical patterns, interpretive framings, and genre conventions beyond the training corpus (zero-shot tests). This criterion evaluates the extent to which the model has internalized not merely lexical co-occurrences but the higher-order discursive constraints governing their deployment. Metrics include: thesis variation rate across domains, typology of logical and factual errors, contextual gains (RAG vs. native knowledge), and stylistic synchronicity with target genres. Existing evaluation frameworks — HELM (Liang et al., 2022), BIG-Bench (Srivastava et al., 2023), MMLU (Hendrycks et al., 2021) — assess linguistic competence at the token or task level. They do not evaluate at the discursive level we have defined: the capacity to reproduce genre conventions, enunciative positions, argumentative framings, and actantial structures as these emerge from situated documentary corpora[3].

**C2 — Attributable rational coherence.** The capacity to exhibit enunciative coherence evaluable through double-blind expert judgment, measuring the stability of enunciative positions under situated dialogical challenge. This criterion operationalizes the intentional stance: the question is not whether the system "truly reasons" but whether its outputs sustain a coherent argumentative trajectory that qualified judges cannot distinguish from human-produced reasoning — and, critically, at what point and under what conditions this coherence degrades. Tests include: disciplinary jury evaluations scoring construction of reasoning, argumentative constancy, teleological orientation, ethos, and conversational positioning. Metrics include: goal adaptation, pragmatic coherence in professional outputs, and intra-session stability.

---

[2] The detailed operationalization of these metrics — including instrument design, scoring protocols, inter-rater reliability procedures, and computational pipelines — exceeds the scope of the present article, which aims to establish the conceptual and programmatic framework. A companion paper currently in preparation within the ANACONUM research network will provide the full methodological specification and preliminary empirical results from a pilot study on digital controversy corpora.

[3] The metrics proposed here are designed to fill this gap, extending the enunciative pragmatics framework developed in our prior work on digital controversy analysis (Lakel, 2026).

**C3 — Normative alignment and contextual value compliance.** The capacity for normative alignment through explicit educational procedures conforming to situated ethical norms, operating through the integration of usage charters, documented protocols (RLHF/RLAIF), and community-specific "civility tests." This criterion acknowledges that alignment is not a technical property of the model but a social relation between the model and its community of use — a relation that must be continually negotiated, documented, and audited. Metrics include: appropriate refusal rate, avoided harmfulness index, and adapted civility score.

**C4 — Traceable biography.** The capacity for biographical tracking through successive traceable states — a sine qua non condition for any effective governance permitting the establishment of a genealogy of intended or undergone transformations. Without biographical traceability, no audit is possible, no regression can be identified, and no accountability can be assigned. This criterion draws directly on Latour's principle that the agent is defined by its trajectory through a network of transformations (2005). Tests include: measurement of inter-version deltas, analysis of controlled persistence and forgetting, audit of context and provenance logs, and longitudinal analysis of the same model under constant conditions. Metrics include: inter-version capability differentials, catastrophic forgetting measures, and contextual-to-autonomous memory dependency ratios.

These four criteria establish the perimeter of an evaluation of agency that escapes both technophobic dismissal and anthropomorphic projection, grounded in reproducible protocols administered within situated communities of use.

However, the ontological constraint established in our theoretical framework must be maintained with full rigor. ADAs operate exclusively on documents. They access neither the phenomenal world nor embodied cognition except through symbolic mediation (the documentary sedimentation that constitutes the third level of our ontological sorting). Their semantic layer is a derived artifact whose referential validity depends entirely on the quality, scope, and biases of their source corpora. This configuration entails a precise characterization of their epistemic status: their performance rests not on causal comprehension of the world, the kind of understanding that embodied agents acquire through sensorimotor interaction (Clark, 1997; Varela, Thompson & Rosch, 1991), but on the imitation of discursive patterns historically stabilized in human documentary production. These patterns are not ideologically neutral: they carry the sedimented traces of power relations, institutional norms, genre conventions, and axiological orientations that structured their conditions of production (Foucault, 1969; Pecheux, 1969).

The methodological posture follows directly. We suspend ontological judgment (the undecidable question of whether ADAs "truly understand" or "merely simulate" understanding) in order to concentrate on the empirical evaluation of effective discursive positions through a systematic program of independent public tests designed to situate the agent as speaker. This suspension is not evasive agnosticism. It is a methodological discipline rooted in the pragmatist tradition: from Peirce's insistence that the meaning of a concept lies in its practical consequences (1878), through James's criterion that truth is what works in the way of belief (1907), to Dewey's experimental logic evaluating propositions by their operational outcomes (1938), the pragmatist lineage consistently subordinates ontological claims to observable effects. We apply this principle to ADAs: evaluate what the agent does, document the conditions under which it succeeds and fails, and suspend judgment on what it "is" until such judgment becomes empirically consequential.

### Assessing the socialization of ADAs: four axes of impact across three social scales

The evaluation of ADAs — however rigorous — does not suffice to think their inscription in social arrangements. Evaluation characterizes what these agents can do; it says nothing about what happens when they are deployed at scale within institutions, labor markets, educational systems, and public deliberation. The transformation of technical invention into societal innovation (Alter, 2013; Akrich, Callon & Latour, 2006) requires interrogating the socialization of ADAs across three distinct scales — individual, organizational, and public — and across four axes of impact whose empirical investigation constitutes the third temporal moment of the research program proposed here.

The gap between evaluation and socialization is not merely analytical — it is empirical. The literature on algorithmic control in organizations has documented significant impacts on worker autonomy, professional identity, and managerial authority. As Kellogg, Valentine, and Christin (2020) demonstrate in their systematic review, this research, though substantial, remains fragmented across disciplines (management science, sociology of work, human-computer interaction, critical data studies) and across methods (ethnography, surveys, experiments, platform analysis), producing findings that rarely cumulate into coherent theoretical frameworks. The same fragmentation now characterizes the emerging literature on generative AI in organizations and education, where contradictory findings proliferate without shared evaluation standards, comparable corpora, or even agreed-upon definitions of the phenomena under investigation. The four axes defined below are designed to organize this fragmentation into a cumulative research program whose hypotheses are falsifiable and whose negative results are as informative as positive ones (Raji et al., 2021).

**P1 — Impact on discursive formations and the editorial economy.** The deployment of ADAs at scale reconfigures the conditions under which discourse is produced, circulated, and received — at all three social scales. At the individual level, the delegation of writing to an ADA transforms the relationship between the speaker and her utterance: the enunciative position is no longer occupied by the subject who signs the text but by a hybrid assemblage whose respective contributions — human and artificial — cannot be disentangled from the output alone, undermining the juridical and institutional regimes (copyright attribution, academic authorship, peer review accountability) that presuppose a traceable human source for every published utterance (U.S. Copyright Office, 2023; Thorp, 2023; EU AI Act, 2024). At the organizational level, the integration of ADAs into editorial workflows produces what the ILO (2025) characterizes as a structural transformation of media labor: resource-constrained newsrooms adopt generative AI not as a supplement to human journalism but as a substitute for it, with documented cases of AI-generated content replacing entire editorial teams (G/O Media, NewsCorp), fabricated quotes passing editorial review (Hanson, 2024), and AI-generated stock photography proliferating around geopolitically sensitive events (Oremus & Verma, 2023). The Reuters Institute survey across six countries (Fletcher & Nielsen, 2025) reveals that public expectations regarding AI in journalism remain among the lowest of any sector — lower than in science, healthcare, or education — indicating a persistent deficit of social legitimacy that the pace of organizational adoption systematically ignores. At the level of the public sphere, the discursive consequence is a transformation of the editorial economy itself: when the marginal cost of producing discourse approaches zero, the discursive order is restructured not by the scarcity of expression but by the saturation of utterance. Gillespie (2024), testing three widely available generative AI tools, demonstrates that ADAs tend to reproduce normative identities and narratives, rarely representing non-normative arrangements and perspectives. The hypothesis to be tested is therefore twofold: (a) ADA-assisted discursive production measurably modifies the distribution of genres, enunciative positions, and argumentative framings within the fields where it is deployed; (b) this modification operates in the direction of normative homogenization, reducing the diversity of the discursive order in proportion to the scale of adoption.

**P2 — Impact on truth production and critical reasoning.** The second axis concerns the relationship between ADAs and the social production of veridiction. At the individual level, converging evidence documents a cognitive debt (Kosmyna et al., 2025) — the progressive degradation of cognitive function resulting from sustained delegation of intellectual tasks, corroborated by negative correlations between AI usage and critical thinking (Gerlich, 2025), a confidence-competence paradox where output fluency masks reasoning decline (Lee et al., 2025), and educational studies demonstrating that ADA access optimizes the product while degrading the cognitive process that produces it (Bastani et al., 2025; Budiyono et al., 2025). At the level of the public sphere, human capacity to detect AI-generated content remains close to chance (Heppell et al., 2024) and AI-synthesized images increase susceptibility to false claims (Guo et al., 2025), though the effect is not unidirectional: exposure to AI misinformation can also increase the value attached to credible sources (Campante et al., 2025). The hypothesis is: ADA deployment produces measurable effects on veridiction at individual, organizational, and public scales, varying as a function of the intensity, duration, and modality of use.

**P3 — Impact on scientific production and problem-solving capacity.** The third axis concerns the relationship between ADAs and the social organization of knowledge production. A large-scale analysis of 41.3 million papers (*Nature*, 2025) reveals that AI-augmented researchers publish three times more, receive five times more citations, and advance faster — but this individual acceleration corresponds with a collective contraction of scientific focus, AI-driven science spanning less topical ground and concentrating on existing popular problems rather than exploring new ones. This paradox reproduces, at the scale of global science, the competence compression documented in organizational settings: generative AI substitutes for the absence of expertise rather than augmenting its presence (Brynjolfsson et al., 2023), and induces quality degradation for tasks requiring strategic judgment without signaling its failure to the user (Dell'Acqua et al., 2023). At the frontier of discovery, the evidence is productively contradictory: ADAs accelerate within well-defined solution spaces (AlphaFold, Nobel 2024; Romera-Paredes et al., 2024) but cannot generate original hypotheses or detect experimental anomalies (*Scientific Reports*, 2025). The hypothesis is: ADA effects on knowledge production systematically diverge between incremental optimization (where they accelerate) and paradigmatic innovation (where they may constrain).

**P4 — Impact on normative alignments and intercultural formations.** The fourth axis concerns the relationship between ADAs and the normative orders through which communities organize their collective existence. Evaluations across 107 countries demonstrate that all major LDMs exhibit cultural values resembling English-speaking and Protestant European countries — a systematic Western bias that cultural prompting only partially mitigates (Tao et al., *PNAS Nexus*, 2024) — and that the same model produces culturally divergent responses depending on the language of interaction, confirming the internalization not of a universal discursive competence but of a stratified ensemble of culturally marked formations (Lu et al., *Nature Human Behaviour*, 2025). Generative AI tools systematically overrepresent normative identities and underrepresent non-normative arrangements (Gillespie, 2024), while perceptions of AI themselves vary as a function of cultural identity, collectivist cultures evaluating ADAs through social alignment, individualist cultures through autonomy and transparency (Barnes et al., 2024). When ADAs mediate communication across cultural boundaries, the normative orientations embedded in their corpora risk imposing the discursive norms of the dominant culture while suppressing the enunciative conventions of the subordinate one — reproducing at the discursive level the asymmetries documented at the political and economic levels (Couldry & Mejias, 2019; Choudhury, 2023). The hypothesis is: ADA deployment systematically favors the discursive formations dominant in the training corpora, with measurable effects on normative orientations at individual, organizational, and public scales.

These four axes do not operate in isolation. The contraction of scientific focus (P3) is inseparable from the normative homogenization of discourse (P1); the degradation of critical reasoning (P2) conditions the capacity to detect the cultural biases documented in P4; the editorial economy that restructures the public sphere (P1) determines the informational environment within which truth production (P2) and scientific discovery (P3) operate. The macroeconomic data compound this picture: the American Time Use Survey reveals that workers with high AI exposure work 3 additional hours per week despite efficiency gains, consistent with the organizational capture of productivity dividends long theorized in labor process analysis (Braverman, 1974) — the efficiency does not accrue to the worker but is absorbed through intensified output expectations and compressed deadlines. The digital divide reconfigures along three dimensions: algorithmic literacy (capacity to formulate prompts and evaluate outputs critically), bias criticality (ability to identify systematic distortions in training corpora), and differential access to performant versions, which reproduces socioeconomic stratification through the paywall separating base models from frontier systems (OECD, 2023; Capraro et al., 2024).

Yet the methodological limitations of these studies must be acknowledged with the same rigor applied to their findings — not merely as a conventional caveat but as a diagnostic of the structural deficits that currently prevent the constitution of a cumulative science of ADA socialization. Three deficits are identifiable.

(1) First, the characterization of the LDMs under study remains succinct to the point of analytical vacuity: most studies identify the model by its commercial designation (GPT-4, Claude, Gemini) without documenting the constitutive factors — corpus composition, architectural version, alignment procedures, deployment configuration — that the descriptive nomenclature proposed in the first axis of this program is designed to provide; without this documentation, no result is replicable and no inter-study comparison is warranted.
(2) Second, the evaluation of ADA competences remains confined to technical performance metrics (accuracy, fluency, benchmark scores) rather than to the social functions — discursive generalization, normative alignment, argumentative coherence within situated communities — that the criteria C1–C4 are designed to capture; the consequence is an accumulation of measurements that describe what the model computes but not what it does within the social fields where it operates.
(3) Third, studies of populations and usages remain fragmented across disciplines, methods, sample sizes, and cultural contexts in a manner that permits neither generalization nor cumulation: each study constitutes an isolated observation whose relationship to other observations cannot be established in the absence of shared protocols, comparable corpora, and standardized evaluation instruments.

These three deficits — descriptive, evaluative, and empirical — are not independent; they compound one another, producing a literature that is simultaneously abundant and inconclusive.

The hour has come for an institutionalization and a critical normalization of research on ADA socialization — the construction of shared infrastructures (standardized model documentation, open evaluation batteries, comparable population protocols) without which the convergent but fragile pattern of evidence reviewed above cannot be consolidated into the reliable knowledge base that governance decisions require. What must be maintained with full clarity is that these transformations constitute neither technological fatality nor inevitable progress. They result from design choices and organizational implementation decisions, always singular, always contestable, always made under conditions where the consequences of deployment outpace the capacity to evaluate them (Collingridge, 1980). The fundamental challenge is now precisely delineated: the preservation of cognitive autonomy, epistemic diversity, and normative pluralism under conditions of accelerating algorithmic mediation — and it is not a problem that technology can solve, because it is a problem that technology, in its current mode of deployment, produces.

## CONCLUSION: TOWARD A SOCIAL SCIENCE OF ARTIFICIAL DISCURSIVE AGENTS

This article constitutes a position paper whose ambition is to effect a paradigmatic displacement in the study of large generative models — a displacement grounded in the theoretical traditions of discourse analysis, enunciative pragmatics, and the social sciences, and directed against the twin impasses that currently structure the field: the computational reduction of these systems to statistical language processors, and the speculative projection of existential scenarios onto agents whose actual discursive behaviors remain largely uncharacterized.

(1) The first operation of this displacement consists of reconnecting the analysis of generative AI with the intellectual heritage that the dominant frameworks — computational linguistics, benchmark-driven evaluation, alignment engineering — have systematically bypassed. The theoretical architecture developed here draws on a lineage that runs from Foucault's archaeology of discursive formations (1969) through Pêcheux's materialist discourse analysis (1969, 1975), Bakhtine's theory of speech genres (1984), and the enunciative pragmatics of Benveniste (1966), Ducrot (1984), and Maingueneau (2002), articulated with the speech act tradition (Austin, 1962; Searle, 1969) and extended by the sociology of translation (Akrich, Callon & Latour, 2006) and the pragmatist philosophy of Peirce (1878), James (1907), and Dewey (1938). This is not a decorative appeal to continental philosophy. It is the mobilization of a body of conceptual instruments — discursive formations, enunciative positions, genre constraints, communicative contracts, procedures of discourse control — without which the object under investigation cannot be constituted at the appropriate analytical level. The category "Large Language Model" misidentifies what the training corpus contains because it reads the corpus through an epistemology of language rather than through a theory of discourse. The terminological displacement from LLM to Large Discourse Model is not a relabeling; it is the consequence of an ontological sorting that distinguishes three regulatory instances — phenomenal, cognitive, and discursive — and demonstrates that these systems operate exclusively at the third level, accessing the first two only through documentary mediation. This sorting is what makes it possible to characterize what these models formalize (the discursive conditions under which a documented civilization produces interpretable utterances) without either reducing them to stochastic parrots or inflating them into artificial general intelligences.

(2) The second operation extends the displacement from theoretical characterization to empirical evaluation and social investigation. The category of Artificial Discursive Agent, grounded in the anti-essentialist problematic of the intentional stance (Dennett, 2009) and the symmetrical treatment of human and non-human agents within sociotechnical assemblages (Latour, 2005), defines these systems not by what they "are" but by what they demonstrably do — and proposes the instruments through which what they do can be evaluated (four cumulative criteria: C1–C4), their developmental trajectories documented (four evolutionary axes: A1–A4), and their socialization effects measured across three scales (individual, organizational, public sphere) and four axes of impact (P1–P4). The framework thus articulates three temporally ordered operations — characterize, evaluate, study in society — that constitute the minimal structure of any empirical program capable of producing the knowledge that governance decisions require.

This article does not present empirical results. It establishes the conceptual and programmatic architecture within which such results can be produced, compared, and accumulated. It constitutes, in this respect, a foundational moment — the first step in a research program whose subsequent phases can now be specified. The immediate next step consists of producing exhaustive, systematic literature reviews organized not by the conventional categories of the existing literature (natural language processing, human-computer interaction, AI ethics, platform studies) but by the analytical grid proposed here. The current literature on generative AI is dispersed across disciplines that do not share evaluation standards, comparable corpora, or even agreed-upon definitions of the phenomena under investigation (Kellogg, Valentine & Christin, 2020). The state of the art must be reconstructed along three axes that correspond to the three operations of the program: the characterization of LDMs (what is known about their constitutive factors across the four evolutionary axes A1–A4, and what remains opaque); the evaluation of ADA competences (what existing benchmarks measure and what they fail to capture at the discursive level defined by criteria C1–C4); and the socialization of ADAs (what the empirical evidence reveals about the four axes of impact P1–P4, what contradictions remain unresolved, and what methodological deficits prevent cumulation). These three literature reviews, each organized around the categories that this article proposes, would accomplish two functions simultaneously: mapping the existing knowledge base onto a coherent analytical structure, and identifying with precision the lacunae where new empirical investigation is required.

The second subsequent phase consists of operationalizing the evaluation framework into research norms for ADA studies — standardized protocols whose adoption across research groups would render results comparable and cumulative. Three deficits have been diagnosed in the current literature: the characterization of models under study remains succinct to the point of analytical vacuity; the evaluation of competences remains confined to technical performance metrics rather than social functions; and studies of populations and usages remain fragmented across methods, samples, and contexts that permit neither generalization nor accumulation. The response to these deficits is not a call for methodological uniformity but the construction of a shared descriptive infrastructure: standardized model documentation protocols (specifying, at minimum, the behavioral phenotype of the agent under investigation — dated version, observable characteristics, deployment configuration, accessible context window); evaluation batteries designed to assess discursive functions rather than computational performance alone; and comparable population protocols that specify the situated communities within which ADA competences are tested and socialization effects measured. The analogy with the foundational moment of the social sciences is structural, not rhetorical: when Quetelet (1835) undertook his social physics and Durkheim (1897) sought to correlate social variables, the statistical infrastructure necessary to describe populations did not pre-exist — it had to be constructed through institutional agreements on what would count as a datum (Desrosieres, 1993). The social science of artificial discursive agents finds itself in a comparable situation, and the resolution is the same: construct the infrastructure first, because without it no individual study, however rigorous, can contribute to cumulative knowledge.

The expected contributions of this program operate at two distinct levels whose articulation defines its specificity. At the level of research quality, the adoption of the framework proposed here would address the three compounding deficits identified in the current literature by imposing descriptive, evaluative, and empirical standards that are currently absent. The consequence is not merely an improvement in individual study quality but the constitution of a cumulative evidence base: a body of knowledge in which each new result extends, refines, or contradicts previous results within a common analytical structure, rather than adding another isolated data point to a literature that grows in volume without growing in coherence.

At the level of societal knowledge, this cumulative evidence base would for the first time enable a systematic, empirically grounded comparison of the effects of ADA deployment across the institutional domains — editorial production, education, scientific research, public deliberation, intercultural communication — where the digital transition of societies is now being decided. The four axes of impact (P1–P4) are designed so that results obtained in one domain can be compared with results obtained in another, precisely because they are evaluated through the same criteria and documented through the same protocols. What emerges is not a collection of disciplinary findings but a coordinated cartography of how artificial discursive agents are transforming the conditions under which human societies produce, circulate, and evaluate discourse — the fundamental operation through which social order is constituted and contested.

This cartography provides the empirical foundation — currently absent — for the critical and political reflection that the present moment demands. Democratic societies have previously confronted innovations posing risks to human populations and have developed, through protracted political struggle, the institutional architectures capable of governing them: pharmaceutical regulation, nuclear governance, biosafety protocols (Jasanoff, 2005; Callon, Lascoumes & Barthe, 2001). What distinguishes the present situation is the nature of the risk. It is not physical but cognitive. The threat posed by the unregulated deployment of artificial discursive agents is not that they will destroy infrastructure or poison bodies but that they will progressively erode the cognitive capacities — critical thinking, sustained reasoning, autonomous judgment, epistemic diversity — upon which the very possibility of democratic governance depends (Kosmyna et al., 2025; Gerlich, 2025; Bastani et al., 2025). The convergent evidence reviewed in this article, fragile and methodologically limited though it remains, delineates this risk with sufficient precision to establish that the question is no longer whether governance is necessary but whether the institutional capacity to govern exists — and whether it can be constructed in time. The research program proposed here is designed to contribute to this construction. Not by prescribing governance solutions from a position of theoretical authority, but by producing the empirical knowledge without which governance decisions are made in the dark.

LAKEL Amar
Bordeaux Montaigne University

**Biographical note**

Amar Lakel is Associate Professor (Maître de Conférences HDR) in Information and Communication Sciences at the University of Bordeaux Montaigne. He is a member of the E3D research group within the MICA laboratory (Mediations, Information, Communication, Arts) and coordinates the ANACONUM research network (Analysis of Digital Controversies). His research focuses on the computational analysis of digital controversies, the epistemology of algorithmic interpretation, and the governance of digital transition in democratic societies.